\def\year{2020}\relax
\documentclass[letterpaper]{article} 
\usepackage{aaai20}  
\usepackage{times}  
\usepackage{helvet} 
\usepackage{courier}  
\usepackage[hyphens]{url}  
\usepackage{graphicx} 
\usepackage{graphicx}  
\usepackage{booktabs}
\usepackage{xspace}
\usepackage{multirow}
\usepackage[dvipsnames]{xcolor}

\newcommand{\dataset}{\textsc{MQR}\xspace}

\newcommand{\citet}[1]{\citeauthor{#1} \shortcite{#1}}
\newcommand{\citep}{\cite}
\newcommand{\citealp}[1]{\citeauthor{#1} \citeyear{#1}}

\newsavebox\FrameBox

\def\wa{\textsuperscript{\rm 1}}
\def\wb{\textsuperscript{\rm 2}}
\def\wc{\textsuperscript{\rm 3}}

\title{How to Ask Better Questions? \\
A Large-Scale Multi-Domain Dataset for Rewriting Ill-Formed Questions}
\author{}

\author{Zewei Chu,\wa \thanks{Work performed during internship at Google.} \xspace \Large \textbf{
Mingda Chen},\wb$^*$\thanks{Contributed equally.} \xspace \textbf{Jing Chen},\wc$^\dagger$ \xspace \textbf{Miaosen Wang},\wc$^\dagger$ \\
\Large \textbf{Kevin Gimpel},\wb \xspace \textbf{Manaal Faruqui},\wc \xspace \textbf{Xiance Si}\wc \\ 
\wa The University of Chicago, 5730 S Ellis Ave, Chicago, IL 60637, USA\\
\wb Toyota Technological Institute at Chicago, 6045 S Kenwood Ave, Chicago, IL 60637, USA\\
\wc Google Assistant, 1600 Amphitheatre Pkwy, Mountain View, CA 94043, USA\\
zeweichu@uchicago.edu, \{mchen, kgimpel\}@ttic.edu \\ \{chenjin, miaosen, mfaruqui, sxc\}@google.com
}

\begin{document}

\maketitle

\begin{abstract}

We present a large-scale dataset for the task of rewriting an ill-formed natural language question to a well-formed one. 
Our multi-domain question rewriting (\dataset) dataset is constructed from human contributed Stack Exchange question edit histories. 
The dataset contains 427,719 question pairs which come from 303 domains. 
We provide human annotations for a subset of the dataset as a quality estimate. 
When moving from ill-formed to well-formed questions, the question quality improves by an average of 45 points across three aspects. 
We train sequence-to-sequence neural models on the constructed dataset and obtain an improvement of 13.2\% in BLEU-4 over baseline methods built from other data resources. We release the MQR dataset to encourage research on the problem of question rewriting.\footnote{\url{https://github.com/ZeweiChu/MQR}}
 
\end{abstract}

\section{Introduction}
\label{sec:intro}

\begin{table*}[t]
\setlength{\tabcolsep}{5pt}
\small
\begin{center}
\begin{tabular}{llc}
\toprule
Ill-formed & Well-formed & Category \\
\midrule
Spaghetti carbonara, mixing & How to mix a spaghetti carbonara? & cooking \\
Ethical Investing... where to begin? & How to begin ethical investing? & money \\
charging canon sx 700 battery through powerbank & Can I charge a Canon SX 700 battery using a mobile powerbank? & photo \\
H1B Visa consulate interview timeline & What is the timeline for an H1B visa consulate interview? & expatriates \\
Hanging weight from drywall ceiling & How much weight can I hang from a drywall ceiling? & diy \\
\bottomrule 
\end{tabular}
\end{center}
\caption{\label{tab:example_mqr} Examples of pairs of ill-formed and well-formed questions from the \dataset dataset.}
\end{table*}

Understanding text and voice questions from users is a difficult task as it involves dealing with ``word salad'' and ill-formed text.
Ill-formed questions may arise from imperfect speech recognition systems, search engines, dialogue histories, inputs from low bandwidth devices such as mobile phones, or second language learners, among other sources. 
However, most downstream applications involving questions, such as question answering and semantic parsing, are trained on well-formed natural language. 
In this work, we focus on rewriting textual ill-formed questions, which could improve the performance of such downstream applications.

\citet{faruqui2018identifying} introduced the task of identifying well-formed natural language questions. 
In this paper, we take a step further to investigate methods to rewrite ill-formed questions into well-formed ones without changing their semantics. 
We create a multi-domain question rewriting dataset (\dataset) from human contributed Stack Exchange question edit histories.\footnote{\url{https://archive.org/download/stackexchange}}
This dataset provides pairs of questions: the original ill-formed question and a well-formed question rewritten by the author or community contributors. The dataset contains 427,719 question pairs which come from 303 domains. 
The \dataset dataset is further split into TRAIN and DEV/TEST, where question pairs in DEV/TEST have less $n$-gram overlap but better semantic preservation after rewriting. 
Table~\ref{tab:example_mqr} shows some example question pairs from the \dataset DEV split. 

Our dataset enables us to train models directly for the task of question rewriting. We train neural generation models on our dataset, including 
Long-Short Term Memory networks~(LSTM;~\citealp{Hochreiter:1997:LSM:1246443.1246450})
with attention~\cite{luong-etal-2015-effective} and transformers~\cite{vaswani2017attention}. We show that these models consistently improve the well-formedness of questions although sometimes at the expense of 
semantic drift. 
We compare to approaches that do not use our training dataset, including general-purpose sentence paraphrasing, grammatical error correction (GEC) systems, and round trip neural machine translation.  Methods trained on our dataset greatly outperform those developed from other resources. 
Augmenting our training set with additional question pairs such as Quora or Paralex question pairs \cite{fader2013paraphrase} has mixed impact on this task. 
Our findings from the benchmarked methods suggest potential research directions to improve question quality. 

To summarize our contributions: 
\begin{itemize}
    \item We propose the task of question rewriting: converting textual ill-formed questions to well-formed ones while preserving their semantics. 
    \item We construct a large-scale multi-domain question rewriting dataset \dataset from human generated Stack Exchange question edit histories. The development and test sets are of high quality according to human annotation. 
    The training set is of large-scale. We release the \dataset dataset to encourage research on the question rewriting task. 
    \item 
    We benchmark a variety of neural models trained on the \dataset dataset, neural models trained with other question rewriting datasets, and other paraphrasing techniques. We find that models trained on the \dataset and Quora datasets combined followed by grammatical error correction perform the best in the \dataset question rewriting task. 
\end{itemize}

\section{Related Work}

\subsection{Query and Question Rewriting}
Methods have been developed to reformulate or expand search queries~\cite{jones2006generating}. 
Sometimes query rewriting is performed for sponsored search~\cite{zhang2007query,zhangcomparing}. 
This work differs from our goal as we rewrite ill-formed questions to be well-formed. 

Some work rewrites queries by searching through a database of query logs to find a semantically similar query to replace the original query. 
\citet{de2010learning} compute query similarities for query ranking based on user click information. \citet{dong-etal-2017-learning} learn paraphrases of questions to improve question answering systems. 
\citet{kumar2018translating} translate queries from search engines into natural language questions. They used Bing's search logs and their corresponding clicked question page as a query-to-question dataset. We work on question rewriting without any database of question logs. 

Actively rewriting questions with reinforcement learning has been shown to improve QA systems \cite{buck2017ask}. This work proposes to rewrite questions to fulfill more general quality criteria. 


\subsection{Paraphrase Generation}

A variety of paraphrase generation techniques have been proposed and studied~\cite{barzilay2003learning,bannard2005paraphrasing,androutsopoulos2010survey,madnani-dorr-2010-generating,malakasiotis2011generate,li2019decomposable}.  Recently, \citet{gupta2018deep} use a variational autoencoder to generate paraphrases from sentences and \citet{li2018paraphrase} use deep reinforcement learning to generate paraphrases.  Several have generated paraphrases by separately modeling syntax and semantics~\cite{iyyer2018adversarial,chen-etal-2019-controllable}.

Paraphrase generation has been used in several applications. \citet{cho-etal-2019-paraphrase} use paraphrase generation as a data augmentation technique for natural language understanding. \citet{iyyer2018adversarial} and  \citet{ribeiro2018semantically} generate adversarial paraphrases with surface form variations to measure and improve model robustness. 
\citet{wieting-gimpel-2018-paranmt} generate paraphrases using machine translation on parallel text and use the resulting sentential paraphrase pairs to learn sentence embeddings for semantic textual similarity. 

Our work focuses on question rewriting to improve question qualities, which is different from general sentence paraphrasing. 

\subsection{Text Normalization}
Text normalization~\cite{sproat2001normalization} is the task of converting non-canonical language to ``standard'' writing. Non-canonical language frequently appears in informal domains such as social media postings or other conversational text, user-generated content, such as search queries or product reviews, speech transcriptions, and low-bandwidth input settings such as those found with mobile devices. 
Text normalization is difficult to define precisely and therefore difficult to provide gold standard annotations and evaluate systems for~\cite{eisenstein-2013-bad}. In our setting, rewriting questions is defined implicitly through the choices made by the Stack Exchange community with the goals of helpfulness, clarity, and utility.

\section{Task Definition: Question Rewriting}
\label{sec:questionrewrite}

Given a question $q_i$, potentially ill-formed, the question rewriting task is to convert it to a well-formed natural language question $q_w$ while preserving its semantics and intention.
Following \citet{faruqui2018identifying}, we define a well-formed question as one satisfying the following constraints: 
\begin{itemize}
    \item The question is grammatically correct. Common grammatical errors include misuse of third person singular or verb tense. 
    \item The question does not contain spelling errors. Spelling errors refer specifically to typos and other misspellings, but not to grammatical errors such as third person singular or tense misuse in verbs. 
    \item The question is explicit. A well-formed question must be explicit and end with a question mark. A command or search query-like fragment is not well-formed. 
\end{itemize}

\begin{table*}[t]
\setlength{\tabcolsep}{6pt}
\small
\begin{center}
\begin{tabular}{lcccc}
\toprule
Question & Spelling & Grammar & Explicit & Remark \\
\midrule
How to remove water-based paint? &	1&	1 &	1 	&\\
how can I make quark the music player to work in unity, natty?&	0&	0 &	1 &\\
What is the value to checking in broken unit tests? &	1 &	0 & 1 & to $\rightarrow$ of \\
No room for RO drain saddle? & 1 & 1 & 0 & \\
\bottomrule 
\end{tabular}
\end{center}
\caption{\label{tab:annotation_quality}Examples given to annotators for binary question quality scores. }
\end{table*}

\begin{table*}[t]
\setlength{\tabcolsep}{6pt}
\small
\begin{center}
\begin{tabular}{llc}
\toprule
Question 1 & Question 2 & Equivalent \\ 
\midrule
How to add a lightbox?	& How to add a lightbox to class mix?	& 0 \\ 
how to get md5sum of a string directly in terminal	& How to get the MD5 hash of a string directly in the terminal?  & 1 \\ 
\bottomrule 
\end{tabular}
\end{center}
\caption{\label{tab:annotation_semantics} Example question pairs given to annotators to judge semantic equivalence.}
\end{table*}

\begin{table*}[t]
\setlength{\tabcolsep}{7pt}
\small
\begin{center}
\begin{tabular}{ccc|ccc|c}
\toprule
 \multicolumn{6}{c|}{\bf Quality} & \multirow{2}{*}{\bf Semantic} \\
 \multicolumn{3}{c}{Ill-formed} & \multicolumn{3}{|c|}{Well-formed} & \multirow{2}{*}{\bf Equivalence} \\
 Spelling & Grammar & Explicit & Spelling & Grammar & Explicit & \\
\midrule
139/310=0.45 & 166/310=0.54 & 175/310=0.56 & 282/290=0.97 & 271/290=0.93 & 286/290=0.99 & 183/200=0.92\\
\bottomrule 
\end{tabular}
\end{center}
\caption{\label{tab:dataset_quality}Summary of manual annotations for instances sampled from the DEV and TEST portions of the \dataset dataset.  ``Quality'' are the average quality scores, broken down into three aspects.
``Semantic Equivalence'' is the percentage of question pairs in which the ill-formed and well-formed questions are semantically equivalent. 
The scores are averages of binary scores across both annotators. 
}
\end{table*}

\begin{table}[t]
\setlength{\tabcolsep}{5pt}
\small
\begin{center}
\begin{tabular}{l|c|c}
\toprule
& TRAIN & DEVTEST \\
\midrule
\midrule
\# Categories & 303 & 166 \\
Mean \# instances per category & 320.4 & 25.5 \\
Std \# instances per category & 754.8 & 47.1 \\
Min \# instances per category & 1 & 1 \\
Max \# instances per category & 6237 & 295 \\
\bottomrule 
\end{tabular}
\end{center}
\caption{\label{tab:dataset_categories}Statistics of question pairs (``instances'') from Stack Exchange categories in the \dataset dataset. }
\end{table}

\section{\dataset Dataset Construction and Analysis}
\label{sec:stack_exchange}



We construct our Multi-Domain Question Rewriting (\dataset) dataset from human contributed Stack Exchange question edit histories. 
Stack Exchange is a question answering platform where users post and answer questions as a community. Stack Exchange has its own standard of good questions,\footnote{\url{https://meta.stackexchange.com/questions/92074/what-can-i-do-when-getting-this-question-body-does-not-meet-our-quality-standar}} and their standard aligns well with our definition of well-formed questions. If questions on Stack Exchange do not meet their quality standards, members of the community often volunteer to edit the questions. Such edits typically correct spelling and grammatical errors while making the question more explicit and easier to understand. 

We use 303 sub areas from Stack Exchange data dumps.\footnote{\url{https://archive.org/download/stackexchange}} 
The full list of area names is in the appendix. We do not include Stack Overflow because it is too specific to programming related questions. We also exclude all questions under the following language sub areas: \emph{Chinese}, \emph{German}, \emph{Spanish}, \emph{Russian}, \emph{Japanese}, \emph{Korean}, \emph{Latin}, \emph{Ukrainian}. This ensures that the questions in \dataset are mostly English sentences. Having questions from 303 Stack Exchange sites makes the \dataset dataset cover a broad range of domains. 

We use ``PostHistory.xml'' and ``Posts.xml'' tables of each Stack Exchange site data dump. If a question appears in both ``PostHistory.xml'' and ``Posts.xml'', it means the question was modified. 
We treat the most up-to-date Stack Exchange questions as a well formed-question and treat its version from ``PostHistory.xml'' as ill-formed. 
``PostHistory.xml'' only keeps one edit for each question, so the \dataset dataset does not contain duplicated questions.

The questions in the Stack Exchange raw data dumps do not always fulfill our data quality requirements. For example, some questions after rewriting are still not explicit. Sometimes rewriting introduces or deletes new information and cannot be done correctly without more context or the question description. We thus perform the following steps to filter the question pairs: 
\begin{enumerate}
\item All well-formed questions in the pairs must start with 
``how'', ``why'', ``when'', ``what'', ``which'', ``who'', ``whose'', ``do'', ``where'', ``does'', ``is'', ``are'', ``must'', ``may'', ``need'', ``did'', ``was'', ``were'', ``can'', ``has'', ``have'', ``are''. This step is performed to make sure the questions are explicit questions but not statements or commands. 
\item  To ensure there are no sentences written in non-English languages, we keep questions that contain 80\% or more of valid English characters, including punctuation.\footnote{The list of valid characters after lowercasing is: 0123456789abcdefghijklmnopqrstuvwxyz . , / ? : ; ' \lbrack \rbrack \_ + - = ~ ! @ \# \$ \% \^ \& * ( ) $\vert$ \{ \} $< >$ ` " ' '' and space}
\end{enumerate}
\noindent This yields the \dataset dataset. We use the following heuristic criteria to split \dataset into TRAIN, DEV, and TEST sets: 
\begin{enumerate}
\item  The BLEU scores between well-formed and ill-formed questions (excluding punctuation) are lower than 0.3 in DEV and TEST to ensure large variations after rewriting. 
\item The lists of verbs and nouns between well-formed and ill-formed questions have a Jaccard similarity greater than 0.8 in DEV and TEST. We split DEV and TEST randomly and equally. This yields 2,112 instances in DEV and 2,113 instances in TEST. 
\item  The rest of the question edit pairs (423,495 instances) are placed in the TRAIN set. 
\end{enumerate}

\noindent Examples are shown in Table~\ref{tab:example_mqr}. We release our TRAIN/DEV/TEST splits of the \dataset dataset to encourage research in question rewriting.

\subsection{Dataset Quality}
\label{subsec:dataset_quality}


To understand the quality of the question rewriting examples in the \dataset dataset, we ask human annotators to judge the quality of the questions in the DEV and TEST splits (abbreviated as DEVTEST onward). Specifically, we take both ill-formed and well-formed questions in DEVTEST 
and ask human annotators to annotate the following three aspects regarding each question~\cite{faruqui2018identifying}: 
\begin{enumerate}
    \item Is the question grammatically correct?
    \item Is the spelling correct? Misuse of third person singular or past tense in verbs are considered grammatical errors instead of spelling errors. Missing question mark in the end of a question is also considered as spelling errors. 
    \item Is the question an explicit question, rather than a search query, a command, or a statement? 
\end{enumerate}
The annotators were asked to annotate each aspect with a binary (0/1) answer.
Examples of questions provided to the annotators are in Table~\ref{tab:annotation_quality}. 
We consider all ``How to'' questions (``How to unlock GT90 in Gran Turismo 2?'') as grammatical. Although it is not a complete sentence, this kind of question is quite common in our dataset and therefore we choose to treat it as grammatically correct.

The ill-formed and well-formed questions are shuffled so the annotators do not have any prior knowledge or bias regarding these questions during annotation. 
We randomly sample 300 questions from the shuffled DEVTEST questions, among which 145 examples are 
well-formed and 155 are ill-formed. 
Two annotators produce a judgment for each of the three aspects for all 300 questions.

The above annotation task considers a single question at a time. We also consider an annotation task related to the quality of a question \emph{pair}, specifically whether the two questions in the pair are semantically equivalent.
If rewriting introduces additional information, then the question rewriting task may require additional context to be performed, even for a human writer. 
This may happen when a user changes the question content or the question title is modified based on the additional description about the question. 
In the \dataset dataset, we focus on question rewriting tasks that can be performed without extra information. 

We randomly sample 100 question pairs from DEVTEST for annotation of semantic equivalence. Two annotators produced binary judgments for all 100 pairs. 
Example pairs are shown in Table~\ref{tab:annotation_semantics}.

Table~\ref{tab:dataset_quality} summarizes the human annotations of the quality of the DEVTEST portion of the \dataset dataset. 
We summed up the binary scores from two annotators. 
There are clear differences between ill-formed and well-formed questions. Ill-formed question are indeed ill-formed and well-formed questions are generally of high quality.
The average score over three aspects improves by 45 points from ill-formed to well-formed questions. 
Over 90\% of the question pairs possess semantic equivalence, i.e., they do not introduce or delete information. Therefore, the vast majority of rewrites can be performed without extra information. 

The Cohen's Kappa inter-rater reliability scores~\cite{mchugh2012interrater} are 0.83, 0.77, and 0.89 respectively for the question quality annotations, and 0.86 for question semantic equivalence. These values show good inter-rater agreement on the annotations of the qualities and semantic equivalences of the \dataset question pairs. 

\subsection{Dataset Domains}

As the \dataset dataset is constructed from 303 sub areas of the Stack Exchange networks, it covers a wide range of question domains. Table~\ref{tab:dataset_categories} summarizes the number of categories in the TRAIN and DEVTEST portions of the \dataset dataset, as well as the mean, standard deviation, minimum, and maximum number of instances per categories. 

The number of questions from each sub area is not evenly distributed due to the fact that some sub areas are more popular and have more questions than the others, but the DEV/TEST splits still cover a reasonably large range of domains.

The most common categories in DEV and TEST are ``diy''(295), ``askubuntu''(288), ``math''(250), ``gaming''(189), and ``physics''(140). 
The least common categories are mostly ``Meta Stack Exchange'' websites where people ask questions regarding the policies of posting questions on Stack Exchange sites. 
The most common categories in TRAIN are ``askubuntu''(6237), ``math''(5933), ``gaming''(3938), ``diy''(2791), and ``2604''(scifi).

\begin{table*}[t]
\setlength{\tabcolsep}{5pt}
\small
\begin{center}
\begin{tabular}{l|c|ccc|c}
\toprule
& \bf BLEU-4 & \bf ROUGE-1 & \bf ROUGE-2 & \bf ROUGE-L & \bf METEOR \\
\midrule
Ill-formed & 5.9 & 50.9 & 19.4 & 45.5 & 33.4 \\
\midrule
\multicolumn{6}{l}{\bf Models trained on \dataset} \\
\midrule
LSTM seq-to-seq with attention & 19.2 & 55.8 & 28.3 & 52.8 & 32.7 \\
Transformer & \bf 22.1 & \bf 59.8 & \bf 32.2 & \bf 56.6 & \bf 36.4 \\
\midrule
\multicolumn{6}{l}{\bf Methods built from other resources} \\
\midrule
Grammatical error correction & 13.1 & 52.4 & 24.4 & 47.5 & 34.4 \\
Round trip NMT (Pivot: De) & 9.9 & 41.6 & 16.8 & 38.2 & 28.4 \\
Round trip NMT (Pivot: Fr) & 9.3 & 40.4 & 15.7 & 36.9 & 27.5 \\
Paraphrase generator trained on ParaNMT & 4.9 & 24.8 & 7.5 & 21.8 & 18.8 \\
\bottomrule 
\end{tabular}
\end{center}
\caption{\label{tab:experiment_models} Results on \dataset TEST set. 
The ``Ill-formed'' shows metric scores for the questions in TEST without rewriting. 
The next portion shows results for models trained on the TRAIN portion of \dataset. 
The lower portion shows results for methods using other models and/or datasets.
} 
\end{table*}

\begin{table*}[t]
\setlength{\tabcolsep}{5pt}
\small
\begin{center}
\begin{tabular}{l|c|ccc|c}
\toprule
\bf Training Dataset & \bf BLEU-4 & \bf ROUGE-1 & \bf ROUGE-2 & \bf ROUGE-L & \bf METEOR \\
\midrule
\dataset TRAIN & 22.1 & 59.8 & 32.2 & 56.6 & 36.4 \\
\dataset TRAIN + $\langle$well-formed, well-formed$\rangle$ pairs & 21.1 & \bf 61.4 & 32.1 & \bf 58.0 & \bf 36.8\\
\dataset TRAIN + Quora & \bf 23.6 & 60.5 & \bf 33.4 & 57.5 & \bf 36.8 \\
\dataset TRAIN + Paralex & 21.7 & 58.3 & 31.3 & 55.3 & 35.7 \\
\dataset TRAIN + Quora + Paralex & 23.1 & 60.3 & 33.0 & 57.2 & 36.7 \\
\bottomrule 
\end{tabular}
\end{center}
\caption{\label{tab:experiment_datasets} Results showing how additional training data affects performance for the transformer model.}
\end{table*}

\begin{table*}[t]
\setlength{\tabcolsep}{5pt}
\small
\begin{center}
\begin{tabular}{l|c|ccc|c}
\toprule
& \bf BLEU-4 & \bf ROUGE-1 & \bf ROUGE-2 & \bf ROUGE-L & \bf METEOR \\
\midrule
Transformer (\dataset + Quora) & 23.6 & 60.5 & 33.4 & 57.5 & 36.8 \\
GEC & 13.1 & 52.4 & 24.4 & 47.5 & 34.4 \\
GEC $\rightarrow$ Transformer (\dataset + Quora) & 24.8 & 60.2 & 33.9 & 57.3 & 36.8 \\
Transformer (\dataset + Quora) $\rightarrow$ GEC & \bf 26.3 & \bf 61.0 & \bf 35.4 & \bf 58.1 & \bf 37.3 \\
Transformer (\dataset) $\rightarrow$ Transformer (\dataset) 
& 20.4 & 55.8 & 29.2 & 52.5 & 35.1 \\
\bottomrule 
\end{tabular}
\end{center}
\caption{\label{tab:experiment_ensemble} Methods combining transformer trained on \dataset + Quora with GEC. ``A $\rightarrow$ B'' means running method A followed by method B on method A's output.}
\end{table*}

\begin{table*}[t]
\setlength{\tabcolsep}{5pt}
\small
\begin{center}
\begin{tabular}{l|ccc|cc}
\toprule
& Spelling & Grammar & Explicit & Semantics wrt. ill-formed & Semantics wrt. well-formed \\
\midrule
Ill formed & 0.31 & 0.41 & 0.61 & - & - \\
GEC & 0.39 & 0.56 & 0.59 & 1.00 & 0.84 \\
Transformer (\dataset + Quora) & 0.96 & 0.75 & 1.00 &  0.67 & 0.56 \\
Transformer (\dataset + Quora) $\rightarrow$ GEC & 0.96 & 0.91 & 1.00 & 0.71 & 0.63 \\
\bottomrule 
\end{tabular}
\end{center}
\caption{\label{tab:human_evaluation} Results of human evaluation of three models on 75 test examples. 
}
\end{table*}

\begin{table*}[t]
\setlength{\tabcolsep}{5pt}
\small
\begin{center}
\begin{tabular}{l|c|ccc|c}
model & question & S & G & E & Semantics \\
\toprule
Ill-formed & best way of widening butcherblock countertop? & 0 & 0 & 0 & 1 \\
Well-formed & What's the best way to widen a butcherblock countertop? & 1 & 1 & 1 & - \\
Trans. & How can I widen a butcherblock countertop? & 1 & 1 & 1 & 0 \\
LSTM & What is the best way of widening butcherblock countertop? & 1 & 0 & 1 & 1 \\
GEC & best way of widening butcherblock countertop? & 0 & 1 & 0 & 1 \\
Round trip NMT (Pivot: De) & best way to extend the racquet counter pole? & 0 & 1 & 0 & 0\\
Round trip NMT (Pivot: Fr) & What is the best way to expand the butcherblock? & 1 & 1 & 1 & 0？ \\
ParaNMT & the best way to expand the countertop of the butcher ? & 1 & 1？ & 0 & 1？ \\
Trans. (\dataset + Quora) & What is the best way of widebutcherblock countertop? & 0 & 0 & 0 & 0\\
Trans. (\dataset + Quora) $\rightarrow$ GEC & What is the best way of widebitcherblock countertop? & 0 & 0 & 0 & 0 \\
\midrule 
Ill-formed & drawing polygons from python console & 0 & 1 & 0 & 1 \\
Well-formed & How to draw polygons from the python console? & 1 & 1 & 1 & -\\
Trans. & How to draw polygons from a python console? &1&1&1&1\\
LSTM & How can I draw polygons from a Python console? &1 &1&1 &1\\
GEC & drawing polygons from python console &0&0&0&1\\
Round trip NMT (Pivot: De) & Drawing polygons from the Python console &0&1&0&1\\
Round trip NMT (Pivot: Fr) & polygons of the python console &0&1&0&0\\
ParaNMT & drawing polygons from python console & 0 & 0 & 0 & 1\\
Trans. (\dataset + Quora) & How to draw polygons from python console?  & 1 & 0 & 1 & 1 \\
Trans. (\dataset + Quora) $\rightarrow$ GEC & How to draw polygons from a python console? & 1 & 1 & 1 & 1\\
\bottomrule 
\end{tabular}
\end{center}
\caption{\label{tab:example_model_rewrite}Examples of ill-formed question rewritten by models with human annotations. (S = Spelling, G = Grammar, and E = Explicit) The last column shows semantic equivalence with the well-formed questions. }
\end{table*}

\section{Models and Experiments}

In this section, we describe the models and methods we benchmarked to perform the task of question rewriting. 

To evaluate model performance, we apply our trained models to rewrite the ill-formed questions in TEST and treat the well-formed question in each pair 
as the reference sentence. We then compute BLEU-4~\cite{papineni2002bleu}, ROUGE-1, ROUGE-2, ROUGE-L~\cite{linrouge}, and METEOR~\cite{banerjee2005meteor} scores.\footnote{The BLEU-4 and METEOR scores are calculated using \url{https://github.com/Maluuba/nlg-eval}. ROUGE-1, ROUGE-2, and ROUGE-L are calculated using \url{https://github.com/pltrdy/rouge}.} 
As a baseline, we also evaluate the original ill-formed question using the automatic metrics. 

\subsection{Models Trained on \dataset}

\paragraph{Transformer.}

We use the Tensor2Tensor~\cite{tensor2tensor} implementation of the transformer model~\cite{vaswani2017attention}. We use their ``transformer\_base'' hyperparameter setting. The details are as follows: batch size 4096, hidden size 512, 8 attention heads, 6 transformer encoder and decoder layers, learning rate 0.1 and 4000 warm-up steps. 
We train the model for 250,000 steps and perform early stopping using the loss values on the DEV set. 

In following sections, when a transformer model is used, we follow the same setting as described above. 

\paragraph{LSTM Sequence to Sequence Model with Attention.}
We use the attention mechanism proposed by \cite{luong-etal-2015-effective}. We use the Tensor2Tensor implementation~\cite{tensor2tensor} with their provided Luong Attention hyperparameter settings. We set batch size to 4096. The hidden size is 1000 and we use 4 LSTM hidden layers following \cite{luong-etal-2015-effective}. 

\subsection{Methods Built from Other Resources}

We also benchmark other methods involving different training datasets and models. All the methods in this subsection use transformer models.

\paragraph{Round Trip Neural Machine Translation.}
Round trip neural machine translation is an effective approach for question or sentence paraphrasing~\cite{mallinson-etal-2017-paraphrasing,dong-etal-2017-learning,iyyer2018adversarial}. 
It first translates a sentence to another pivot language, then translates it back to the original language. 
We consider the use of both German (De) and French (Fr) as the pivot language, so we require translation systems for En$\leftrightarrow$De and En$\leftrightarrow$Fr. 

The English-German translation models are trained on WMT datasets, including News Commentary 13, Europarl v7, and Common Crawl, and evaluated on newstest2013 for early stopping. On the newstest2013 dev set, the En$\rightarrow$De model reaches a BLEU-4 score of 19.6, and the De$\rightarrow$En model reaches a BLEU-4 score of 24.6. 

The English-French models are trained on Common Crawl 13, Europarl v7, News Commentary v9, Giga release 2, and UN doc 2000. On the newstest2013 dev set, the En$\rightarrow$Fr model reaches a BLEU-4 score of 25.6, and the Fr$\rightarrow$En model reaches a BLEU-4 score of 26.1. 

\paragraph{Grammatical Error Correction (GEC).}
As some ill-formed questions are not grammatical, we benchmark a state-of-the-art grammatical error correction system on this task. 
We use the system of~\cite{lichtarge2019corpora}, a GEC ensemble model trained from Wikipedia edit histories and round trip translations. 


\paragraph{Paraphrase Generator Trained on ParaNMT.}

We also train a paraphrase generation model on a subset of the ParaNMT dataset~\cite{wieting-gimpel-2018-paranmt}, which was created automatically by using neural machine translation to translate the Czech side of a large Czech-English parallel corpus. 
We use the filtered subset of 5M pairs provided by the authors. For each pair of paraphrases (S1 and S2) in the dataset, we train the model to rewrite from S1 to S2 and also rewrite from S2 to S1. We use the \dataset DEV set for early stopping during training.

\subsection{Results}

Table~\ref{tab:experiment_models} shows the performance of the models and methods described above. 
Among these methods models trained on \dataset work best. 
GEC corrects grammatical errors and spelling errors, so it also improves the question quality in  rewriting. 
Round trip neural machine translation is a faithful rewrite of the questions, and it naturally corrects some spelling and grammatical errors during both rounds of translation due to the strong language models present in the NMT models. However, it fails in converting commands and statements into questions.


The paraphrase generator trained on ParaNMT does not perform well, likely because of domain difference (there are not many questions in ParaNMT). It also is unlikely to convert non-question sentences into explicit questions.



\subsection{Additional Training Data}



We consider two additional data resources to improve question rewriting models. 

The first resource is the Quora Question Pairs dataset.\footnote{\url{https://www.quora.com/q/quoradata/First-Quora-Dataset-Release-Question-Pairs}} 
This dataset contains question pairs from Quora, an online question answering community. 
Some question pairs are marked as duplicate by human annotators and other are not. 
We consider all Quora Question Pairs (Q1 and Q2) marked as duplicate as additional training data. We train the model to rewrite from Q1 to Q2 and also from Q2 to Q1. This gives us 298,364 more question pairs for training. 

The second resource is the Paralex dataset~\cite{fader2013paraphrase}. 
The questions in Paralex are scraped from WikiAnswers,\footnote{\url{http://wiki.answers.com/}}  where questions with similar content are clustered. 
As questions in the Paralex dataset may be noisy, we use the annotation from \cite{faruqui2018identifying}. 
Following their standard, we treat all questions with scores higher than 0.8 as well-formed questions. For each well-formed question, we take all questions in the same Paralex question cluster and construct pairs to rewrite from other questions in the cluster to the single well-formed question. 
This gives us 169,682 extra question pairs for training. 

We also tried adding ``identity'' training examples in which the well-formed questions from the \dataset TRAIN set are repeated to form a question pair. 

The results of adding  training data are summarized in Table~\ref{tab:experiment_datasets}. 
Adding the identity pairs improves the ROUGE and METEOR scores, which are focused more on recall, while harming BLEU, which is focused on precision. We hypothesize that adding auto-encoding data improves semantic preservation, which is expected to help the recall-oriented metrics.  
Adding Quora Question Pairs improves performance on TEST but adding Paralex pairs does not. The reason may stem from domain differences: WikiAnswers (used in Paralex) is focused on factoid questions answered by encyclopedic knowledge while Quora and Stack Exchange questions are mainly answered by community contributors. 
Semantic drift occurs more often in Paralex question pairs as Paralex is constructed from question clusters, and a cluster often contains more than 5 questions with significant variation. 

\subsection{Combining Methods}

In addition to the aforementioned methods, we also try combining multiple approaches.  Table~\ref{tab:experiment_ensemble} shows results when combining GEC and the Quora-augmented transformer model. We find that combining GEC and a transformer question rewriting model achieves better results than each alone. In particular, it is best to first rewrite the question using the transformer trained on \dataset + Quora, then run GEC on the output.

We also tried applying the transformer (trained on \dataset) twice, but it hurts the performance compared to applying it only once (see Table~\ref{tab:experiment_ensemble}).

\subsection{Human Evaluation}

To better evaluate model performance, 
we conduct a human evaluation on the model rewritten questions following the same guidelines from the ``Dataset Quality'' subsection. 
Among the 300 questions annotated earlier, 
we chose the ill-formed questions from the TEST split, which yields 75 questions. 
We evaluate questions rewritten by three methods (Transformer (\dataset + Quora), GEC, and Transformer (\dataset + Quora) $\rightarrow$ GEC), and ask annotators to determine the qualities of the rewritten questions.
To understand if question meanings change after rewriting, we also annotate whether a model rewritten question is semantically equivalent to the ill-formed question or equivalent to the well-formed one.


Table~\ref{tab:human_evaluation} shows the annotations from two annotators.  When the two annotators disagree, a judge makes a final decision. 
Note that the examples annotated here are a subset of those annotated in Table~\ref{tab:dataset_quality}, so the first row is different from the ill-formed questions in Table~\ref{tab:dataset_quality}. 
According to the annotations, the GEC method slightly improves the question quality scores. Although Table~\ref{tab:experiment_models} shows that GEC improves the question quality by some automatic metrics, it simply corrects a few grammatical errors and the rewritten questions still do not meet the standards of human annotators. However, the GEC model is good at preserving question semantics. 

The Transformer (\dataset + Quora) model and Transformer (\dataset + Quora) $\rightarrow$ GEC excel at improving question quality in all three aspects, but they suffer from semantic drift. 
This suggests that future work should focus on solving the problem of semantic drift when building question rewriting models. 

Table~\ref{tab:example_model_rewrite} shows two example questions rewritten by different methods. The questions rewritten by GEC remain unchanged but are still of low quality, whereas ParaNMT and round trip NMT make a variety of changes, resulting in large variations in question quality and semantics. 
Methods trained on \dataset excel at converting ill-formed questions into explicit ones (e.g., adding ``What is'' in the first example and ``How to'' in the second example), but sometimes make grammatical errors (e.g., Trans. (\dataset + Quora) misses ``a'' in the second example). 
According to Table~\ref{tab:experiment_ensemble}, combining neural models trained on \dataset and GEC achieves the best results in automatic metrics. However, they still suffer from semantic drift. 
In the first example of Table~\ref{tab:example_model_rewrite}, the last two rewrites show significant semantic mistakes, generating non-existent words ``widebutcherblock'' and ``widebitcherblock''.

\section{Conclusion and Future Work}

We proposed the task of question rewriting and produced a novel dataset \dataset to target it. Our evaluation shows consistent gains in metric scores when using our dataset compared to systems derived from previous resources. 
A key challenge for future work is to design better models to rewrite ill-formed questions without changing their semantics. 
Alternatively, we could attempt to model the process whereby question content changes. Sometimes community members do change the content of questions in online forums. Such rewrites typically require extra context information, such as the question description. Additional work will be needed to address this context-sensitive question rewriting task. 

\section{ Acknowledgments}
We thank Shankar Kumar, Zi Yang, Yiran Zhang, Rahul Gupta, Dekang Lin, Yuchen Lin, Guan-lin Chao, Llion Jones, and Amarnag Subramanya for their helpful discussions and suggestions.

\bibliography{aaai}
\bibliographystyle{aaai}

\end{document}